\documentclass{INTERSPEECH2023}
\usepackage{multirow}
\usepackage{placeins}
\usepackage[belowskip=-4mm,aboveskip=5pt]{caption}
\usepackage{cite} 

\interspeechcameraready

\title{A Novel Self-training Approach for Low-resource Speech Recognition}
\name{Satwinder Singh, Feng Hou, Ruili Wang$^\ast$}
\address{
  School of Mathematical and Computational Sciences, Massey University, Auckland, New Zealand}
\email{\{s.singh4,f.hou,ruili.wang\}@massey.ac.nz}

\begin{document}

\maketitle

\begin{abstract}
\vspace{-1mm}

In this paper, we propose a self-training approach for automatic speech recognition (ASR) for low-resource settings. While self-training approaches have been extensively developed and evaluated for high-resource languages such as English, their applications to low-resource languages like Punjabi have been limited, despite the language being spoken by millions globally. The scarcity of annotated data has hindered the development of accurate ASR systems, especially for low-resource languages (e.g., Punjabi and Māori languages). To address this issue, we propose an effective self-training approach that generates highly accurate pseudo-labels for unlabeled low-resource speech. Our experimental analysis demonstrates that our approach significantly improves word error rate, achieving a relative improvement of 14.94\% compared to a baseline model across four real speech datasets. Further, our proposed approach reports the best results on the Common Voice Punjabi dataset. 

\end{abstract}
\noindent\textbf{Index Terms}: self-training, low-resource, Punjabi ASR

\section{Introduction}
\vspace{-1mm}
The development of end-to-end (E2E) automatic speech recognition (ASR) systems for high-resource languages has made significant strides in recent years, resulting in an excellent performance \cite{baevski2020wav2vec}. The key factors contributing to this high performance include the presence of large annotated datasets, advancements in deep learning algorithms, and easy access to high-performance computational resources. However, the same cannot be said for ASR systems designed for low-resource languages. These systems still fall short in terms of performance when compared to their high-resource counterparts. The primary reason for this discrepancy is the lack of annotated speech datasets and their availability in the public domain. Without these resources, it is difficult for ASR systems to be trained effectively and perform at the same level as those designed for high-resource languages.
\par
The research community has been actively exploring various ways to address the scarcity of annotated datasets for speech recognition systems \cite{ragni2014data, zhou2018multilingual, yi2020applying, singh2022improved, singh2022enhancing}. The process of annotating highly accurate speech datasets is a labor-intensive task that requires a significant amount of time and resources. To overcome these challenges, many researchers have proposed self-training or pseudo-labeling (PL) based methods \cite{kahn2020self}. These methods leverage the vast amount of available unlabeled data in conjunction with a small labeled dataset to improve the performance of ASR systems. Self-training has gained increasing interest in a wide range of research areas, including computer vision \cite{lee2013pseudo, berthelot2019mixmatch}, natural language processing \cite{her2019evisiting, li2018pseudo}, and more recently, in the field of speech recognition \cite{ragni2014data, park2020improved, xu2020iterative, higuchi2021momentum, moritz2021semi}. These methods have shown promise in addressing the shortage of annotated data and improving the performance of ASR systems.
\par
The fundamental concept behind self-training is to first train an initial seed model, either from scratch or by fine-tuning an existing pre-trained model, using a limited amount of labeled data. This seed model is then used to generate pseudo-labels for unlabeled data. The process can be repeated over multiple iterations in order to improve the quality of the generated pseudo-labels. To ensure that only highly accurate pseudo-labels are used, various filtering methods are applied to filter out any incorrect labels \cite{park2020improved}. Additionally, a robust language model can be utilized to achieve accurate decoding of ASR transcriptions, further improving the quality of the generated pseudo-labels.
\par
Most of the previous and recent work in self-training/ pseudo-labeling is based on and demonstrated on high-resource languages like English \cite{ragni2014data, park2020improved, xu2020iterative, higuchi2021momentum, likhomanenko2020slimipl}. However, most of the languages of the world are low-resource and the ASR system designed for them performs poorly when compared to those with high-resource languages. With that in mind, in this work, we focus to design a self-training approach for one of the most widely spoken languages in the world, i.e., the Punjabi language. Although the Punjabi language is spoken natively across India and Pakistan region, however, Punjabi speakers are spread across the globe, especially in countries such as Canada, the United States, the United Kingdom, Australia, and New Zealand. Despite having more than 100 million speakers, the Punjabi language still lacks mature ASR systems due to a lack of annotated datasets. 
\par
In the paper, we propose a self-training approach to leverage the available large unlabeled audio data. We adopt a pre-trained self-supervised crosslingual wav2vec 2.0 model (XLSR-53) \cite{conneau2021unsupervised} as our seed model to generate pseudo-labels across multiple iterations. The proposed model first fine-tunes the XLSR-53 seed model on limited Punjabi datasets in a supervised fashion with language model decoding. Afterwards, the fine-tuned model is used to produce pseudo-labels for unlabeled audio data. As we are fine-tuning on limited data, it is common to get a mixed bag of pseudo-labels. To sieve out the erroneous pseudo-labels, we propose a length-normalized confidence score. The pseudo-labels with high confidence scores above the certain cut-off threshold are selected to be included into the dataset. Finally, the model is fine-tuned again from scratch using labeled data and highly confident pseudo-labels. This process is repeated over multiple iterations with gradual confidence score filtration to refine the pseudo-labels. Our empirical results demonstrate that our proposed approach is able to achieve significant WER reduction compared with the baseline and also reports the best results on the Common Voice Punjabi dataset. To the best of our knowledge, this work is the first to explore the self-training approach for the Punjabi language on the mix of datasets (public, non-public, and synthesized datasets). 


\section{Related Work}
 Semi-supervised approaches such as self-supervision and self-training have been active areas of research for the past few years. The goal of these approaches is to leverage the large unlabeled speech data to improve the overall accuracy of the ASR systems. These approaches showed significant results for low-resource languages. Self-training is a fairly simple yet effective semi-supervised approach for utilizing unlabeled audio data. Self-training has been applied in computer vision and natural language processing tasks. However, it is recently adopted for E2E speech recognition. 
\par
Kahn et al. \cite{kahn2020self} proposed a self-training approach for E2E ASR model. The proposed system incorporate a pseudo-label filtering method and language model decoding to improve the word error rates. Additionally, an ensemble of multiple models showed further performance improvement. Further, inspired by noisy student training (NST) from the image processing domain, Park et al. \cite{park2020improved} adopted this method for speech recognition. The proposed approach exploited the adaptive SpecAugment \cite{park2019specaugment} augmentation method. The NST approach incorporated filtering, balancing, and mixing techniques to carefully select highly accurate pseudo-labels. Furthermore, rather than selecting pseudo-labels over just one iteration, iterative pseudo-labeling (IPL) \cite{xu2020iterative} refines the acoustic model over multiple iterations. IPL effectively generates the pseudo-labels for a subset of unlabeled data rather than labeling the entire unlabeled data. Also, IPL fine-tuned an existing model instead of training an entirely new model.
\par
Recently, Higuchi et al. \cite{higuchi2021momentum} presented momentum pseudo-labeling (MPL), which incorporated online and offline models. At different time steps, weights of multiple models are averaged using the momentum-based moving average to refine pseudo-labels. The online model (also called the teacher model), hypothesizes the pseudo-labels produced by the offline model (also referred to as the student model) on the go. The MPL approach outperformed standard pseudo-labels and IPL approaches. The extension to this work is presented in \cite{higuchi2022advancing}, where MPL is combined with IPL. The resulting approach improved the seed model by exploiting language model knowledge using the IPL strategy.  Additionally, the Conformer-based ASR architecture further enhanced the overall accuracy of the system. 
\par
While most of the self-training approaches utilized language models for decoding, the SlimIPL \cite{likhomanenko2020slimipl} method performed pseudo-labeling without any language model. The proposed model is built upon the IPL approach, which produced pseudo-labels over multiple iterations using a single model. SlimIPL also incorporated a dynamic cache for better and more stable model training. Furthermore, Lugosch et al. \cite{lugosch2022pseudo} presented a self-training approach to multilingually generate pseudo-labels for 60 languages. The proposed approach first pre-trained a single multilingual model followed by fine-tuning it on the target language using semi-supervision. The fine-tuned model is then used to produce pseudo-labels for that particular language, which in turn is utilized to train the final model. Overall, the multilingual pseudo-labeling demonstrated good performance and better generalization to the data from different domains. 
\par
Xu et al. \cite{xu2021self} presented a study, which showed that self-training and pre-training can effectively be combined and are complementary to each other. Pre-training based self-supervised approaches like wav2vec \cite{baevski2020wav2vec} and its variants demonstrated excellent performance for low-resource languages. They leveraged unlabeled audio data and learn representations using contrastive learning. When combined with self-training, self-supervision showed a better generation of pseudo-labels and achieved significant results. Similarly, Jin et al. \cite{jin2022filter} exploited a self-supervised pre-trained wav2vec 2.0 model for pseudo-labeling. The proposed method filtered out the low-probability pseudo-labels and only selected the pseudo-labels with the highest probability scores. Further, self-training has also been used to improve representation learning in crosslingual settings \cite{zhang2021xlst}. Furthermore, Khurana et al. \cite{khurana2021unsupervised}  proposed the DUST (Dropout-based Uncertainty-driven Self-Training) approach to alleviate the issues caused by domain mismatch. The proposed approach efficiently  rejected pseudo-labels that exhibit high levels of uncertainty.
\par
Until recently, limited work has been carried out in the Punjabi language. Previous efforts relied heavily on statistical and hybrid models, using mostly non-public datasets \cite{dua2012punjabi, guglani2018continuous, kadyan2019comparative, kumar2021autossr}. However, in our previous work \cite{singh2022enhancing}, we demonstrated that E2E models trained with synthetic speech and real speech datasets could enhance the effectiveness of E2E ASR models for Punjabi.

\section{Proposed Approach}
We propose a very simple, yet effective self-training method for the Punjabi language. However, the proposed method could work for any language having a small amount of annotated data and large unlabeled audio data. 

\subsection{Pre-trained seed model}
For the strong baseline seed model, we adopt a self-supervised pre-trained wav2vec 2.0 based crosslingual XLSR-53 model \cite{conneau2021unsupervised}. The XLSR-53 model is pre-trained on 53 languages using a contrastive learning approach to learn audio representations in an unsupervised manner. The crosslingual wav2vec model maps raw input speech to latent speech representations, which are then transformed into a set of discrete representations using a quantization module. These discrete representations are shared across multiple languages (in this case 53 languages excluding the Punjabi language) and used as the target for the self-supervised objective. Conneau et al. \cite{conneau2021unsupervised} showed that XLSR-53 could greatly improve the character error rate (CER) and word error rate (WER) in the case of low-resource languages.
\par
Although pre-trained XLSR-53 does not include Punjabi in the pre-training setup, the model is still able to learn powerful shared feature representation across many languages. The crosslingual nature of the model made it easy to fine-tune the model on any new language with limited labeled data. Following that, rather than pre-training the model using the Punjabi language, we leverage the effective feature representations of the pre-trained model and only fine-tune the model on available Punjabi labeled data. This way, we keep our self-training approach simple and fast. 

\subsection{Self-training approach}
\begin{figure}[!t]
\includegraphics[width=8.2cm]{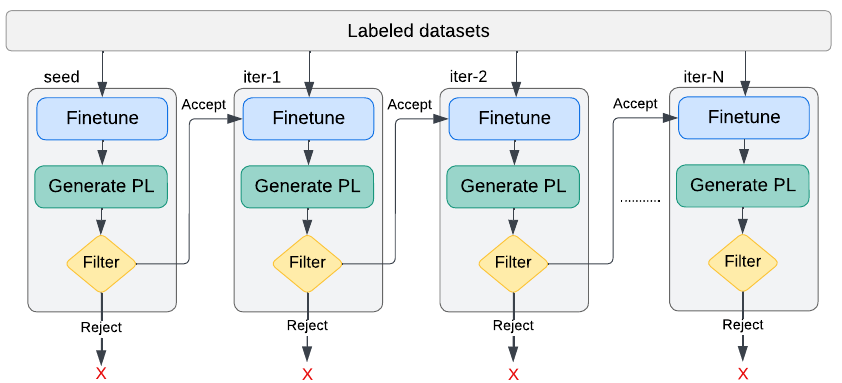}
\caption{Overview of proposed self-training approach.}
\label{fig:pl}
\end{figure}

Our proposed self-training approach is fairly simple as shown in Figure \ref{fig:pl}. Initially, we fine-tune our pre-trained seed acoustic model on available limited labeled Punjabi datasets. Once the model is fine-tuned on the Punjabi language then we use this model to generate the pseudo-labels for unlabeled Punjabi data. The details of labeled and unlabeled datasets are outlined in Section 4.1. At this stage, we also employ the 5-gram KenLM language model \cite{heafield2011kenlm} for decoding the seed model's outputs. 
\par
Since the seed model is only trained on limited Punjabi data, it is fairly common to produce inaccurate pseudo-labels. To tackle the erroneous pseudo-labels, we incorporated confidence based scoring strategy \cite{kahn2020self}. Our language model produces sentence-level shallow fusion scores while decoding. We leverage the shallow fusion score as a confidence score (CS) and normalize it to the length of the sentence. We filter pseudo-labels with different filtration thresholds of confidence scores in each iteration of self-training, i.e., [0.5, 0, -0.5, -1, -1.5, -2, -2.5]. We start with a very strict (0.5) threshold and then with each iteration of self-training, we gradually relax the filtration threshold. Afterwards,  we fine-tune an entirely new model using a combination of labeled and pseudo-labeled datasets. We repeat this whole process over multiple iterations until the ASR model is not improving on the labeled test datasets. 
\par
Our approach is similar to the IPL approach \cite{xu2020iterative}, but with few distinctions. Firstly, we compute pseudo-labels for the entire unlabeled dataset instead of a subset of it. Although it is time-consuming to pseudo-label the entire unlabeled dataset, it would give us more high-quality filtered pseudo-labels for fine-tuning. Secondly, since the XLSR-53 model does not show any improvement over multiple continued fine-tuning phases, we find that fine-tuning a new model from pre-trained XLSR-53 model produces the best results. Therefore, rather than continuing fine-tuning like in IPL, in every iteration, we fine-tune an entirely new model. 

\section{Experimental Setup}
\subsection{Datasets}
To train our seed model we use a range of labeled Punjabi datasets. We utilize four low-resource Punjabi real speech datasets and two synthesized datasets. For real speech datasets, we use the Shrutilipi dataset \cite{bhogale2022effectiveness}, Common Voice (CV) \cite{ardila2020common}, Punjabi Speech \cite{singh2022enhancing}, and 50Languages\footnote{\url{https://www.50languages.com/phrasebook/en/pa}\label{50lang}}. Previously, we have demonstrated the effectiveness of synthetic datasets in improving the performance of the Punjabi ASR system on real speech datasets \cite{singh2022enhancing}. Thus, following that, we use two synthesized datasets i.e. Google-synth and CMU-synth \cite{singh2022enhancing}.
\begin{itemize}
    \item \textbf{Shrutilipi}: The Shrutilipi corpus is a collection of audio and text pairs sourced from public resources such as All India Radio news bulletins, and compiled through data mining techniques. The corpus includes labeled speech data for a total of 12 Indian languages, including Punjabi.
    \item \textbf{Common Voice}: The Common Voice dataset is a crowd-sourced dataset available to the public, featuring speech data in numerous languages. For our purposes, we utilize the Punjabi data made available in the April 2022 release of Common Voice. The dataset comprises of a range of speech samples from a total of 39 male and 12 female speakers, providing a diverse array of recordings.
    \item \textbf{Punjabi Speech}: The Punjabi Speech dataset is a read speech dataset like Common Voice. The dataset consists of speech data from 2 male speakers. 
    \item \textbf{50Languages}: The dataset is extracted from the 50Languages learning platform\textsuperscript{\ref{50lang}}. The dataset comprises speech utterances from a single male speaker.  
    \item \textbf{Google-synth}: The Google-synth dataset is synthesized labeled speech data produced using Google's Cloud text-to-speech (TTS) API\footnote{\url{https://cloud.google.com/text-to-speech}}. The dataset contains speech samples from 2 male and 2 female speakers. 
    \item \textbf{CMU-synth}: The CMU-synth dataset is synthesized using CMU's Clustergen TTS model. The dataset contains speech samples from a single female speaker.
    \item \textbf{Audiobooks}: The Audiobooks dataset consists of speech recordings gathered from open Punjabi audiobooks featured on various YouTube channels. It is an unlabeled dataset used primarily for self-training purposes. The dataset contains audio recordings from multiple speakers and several audiobooks. To facilitate self-training, we preprocess the audio recordings, chunking them down into smaller audio segments no longer than 15 seconds.
\end{itemize}
The synthesized datasets (Google-synth and CMU-synth) are produced using Punjabi text available in the Old Newspaper corpus\footnote{\url{https://www.kaggle.com/datasets/alvations/old-newspapers}}. We ensure that there is no overlap between data used to train the ASR model, self-training, language model, and synthesized datasets.  

\begin{table}[t]
\caption{List of datasets used for our experimentations.}
\centering
\begin{tabular}{lcc}
\toprule
\textbf{Datasets}        & \multicolumn{1}{l}{\textbf{\#Hours}} & \multicolumn{1}{l}{\textbf{\#Utterances}} \\ \toprule\hline
\multicolumn{3}{l}{\textit{Real speech}}                                          \\ \hline
Shrutilipi     & 94                           & 50K                               \\ 
Common Voice   & 1                            & 1210                              \\ 
Punjabi Speech & 4                            & 2431                               \\ 
50Languages    & 3                            & 3955                              \\ \hline
\multicolumn{3}{l}{\textit{Synthesized speech}}                                   \\ \hline
Google-synth   & 38                           & 50K                               \\ 
CMU-synth      & 170                          & 80K                               \\ \hline
\multicolumn{3}{l}{\textit{Unlabeled speech}}                                   \\ \hline
Audiobooks      & 450                          & 322K                               \\ \toprule
\end{tabular}
\label{tab:datasets}
\end{table}

\begin{table*}[t!]
\caption{Experimental results in terms of WER (\%). The best pseudo-label (PL) results are obtained by choosing the best-performing model on labeled datasets over multiple iterations of PL with different confidence thresholds.}
\centering
\begin{tabular}{lcccccc}
\toprule
\multicolumn{1}{c}{\multirow{2}{*}{\textbf{Model}}} & \multicolumn{6}{c}{\textbf{Datasets}}                                                        \\ \cline{2-7} 
\multicolumn{1}{c}{}                       & \textbf{Shrutilipi} & \textbf{Common Voice} & \textbf{Punjabi Speech} & \textbf{50Languages} & \textbf{Google-synth} & \textbf{CMU-synth} \\ \toprule \hline
Seed without LM                           & 17.23      & 29.88        & 18.88           & 28.08       & 13.39        & 6.82    \\ 
Seed with LM (baseline)                            & 16.87      & 14.04        & 9.84           & 24.16       & 2.93         & 4.96      \\ \hline
Raw PL (no filter)                                  & 15.23           & 12.30             & 9.10               &24.00             & 2.75             & 5.12          \\
Best PL                            & \textbf{14.67}          & \textbf{11.42}             & \textbf{8.57}               &\textbf{20.55}             &\textbf{2.69}              &\textbf{4.92}           \\ \toprule
\end{tabular}
\label{tab:results}
\end{table*}

\subsection{Methodology}
For our self-training approach, we adopt a pre-trained wav2vec 2.0 based crosslingual XLSR-53 model. The XLSR-53 model is pre-trained on 53 languages (50K hours) with 300 million parameters. To save time, we only fine-tune the pre-trained model on our labeled datasets instead of pre-training it from scratch. We adopt the pre-trained model by adding a fully connected layer on top of the Transformer. The fully connected layer outputs the characters using Connectionist Temporal Classification (CTC) \cite{graves2006connectionist}. During the fine-tuning, we freeze the weights of the feature encoder. We fine-tune a single XLSR-53 model on combined data from 6 of the labeled datasets as mentioned in Table \ref{tab:datasets}. For all the datasets except Common Voice, we use 8:1:1 splits for train, validation, and test sets. For the Common Voice dataset, we use pre-defined data splits. We fine-tune our model for 10 epochs with a batch size of 32. Further, we optimize the training process using the Adam optimizer \cite{kingma2015adam} with an initial learning rate of 3e-4, which is warmed up for the first 10\% of the updates and afterward, linearly decays during the rest of the updates.  All the experiments are conducted on NVIDIA A100 GPUs and results are reported in terms of word error rate (WER). 
\par
Additionally, we decode the ASR model's output with a 5-gram KenLM language model \cite{heafield2011kenlm}. We train the Punjabi language model using the IndiCorp dataset \cite{kakwani2020indicnlpsuite}. The dataset contains 29.2 million Punjabi sentences (773 million tokens). We fine-tune $\alpha$ and $\beta$ parameters and empirically set $\alpha= 0.7$ and  $\beta= 4.0$. Here $\alpha$ represents the shallow fusion weight for the language model and $\beta$ refers to the weight to adjust the score as per the length of the decoded sequence \cite{kannan2018analysis}.  
Further, for selecting the highly confident pseudo-labels, we adopt a normalized confidence score filtering. 


\vspace{-1mm}
\section{Experimental Results and Discussion}

\subsection{Comparative analysis}
\begin{figure}[ht]
\includegraphics[width=8cm]{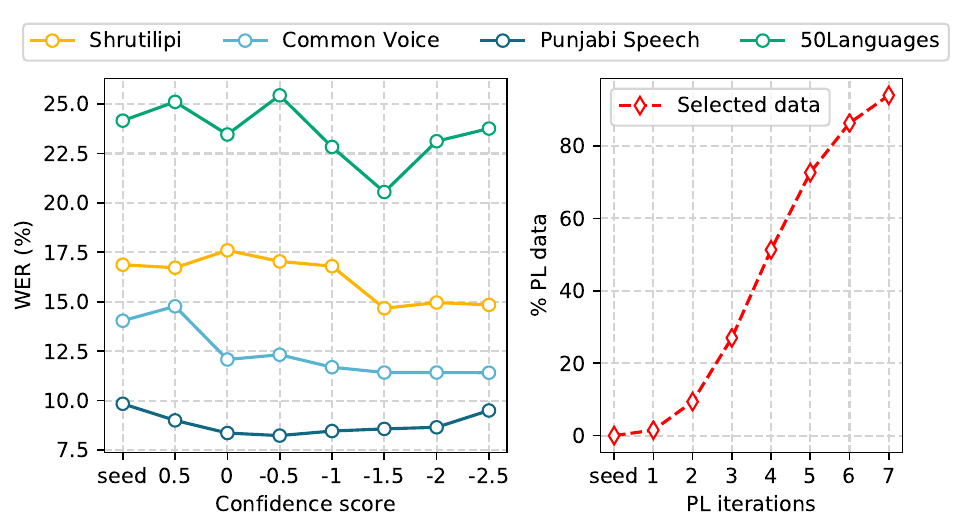}
\caption{Performance on real speech datasets against various confidence score thresholds (left) and \% of selected data over the PL iterations (right).}
\label{fig:wer-cs}
\end{figure}

We report the results in Table \ref{tab:results}. To show the effectiveness of our 5-gram language model (LM), we demonstrate the results on the seed model with and without the language model. We find that our LM decoding further improves the seed model, consequently, we choose seed with LM as our baseline model. All of our experimental results (except seed without LM) are carried out with LM decoding. Raw PL uses all available pseudo-labels, whereas Best PL selects the best-performing pseudo-labels over multiple iterations of pseudo-labeling with different confidence thresholds. The results show that using pseudo-labels with a filter (Best PL) outperforms Raw PL on all datasets, indicating that selecting the best pseudo-label is important to improve the performance of the ASR system. We find that iteration 5 with confidence scores (CS$\ge$-1.5) produces the best results as shown in Table \ref{tab:results} (Best PL) and Figure \ref{fig:wer-cs}.  As compared to the baseline, we find our approach achieves relative improvements of 13.04\%, 18.66\%, 12.91\%, and 14.94\% on real speech Shrutilipi, Common Voice, Punjabi Speech, and 50Languages datasets, respectively. Our PL approach demonstrates the best results on the Common Voice Punjabi dataset with a relative improvement of 50.5\% over previously reported results (23.07 \%  vs 11.42\% WER)\cite{singh2022enhancing}. 
\par
Further, on synthetic datasets (i.e., Google-synth and CMU-synth), our PL approach also achieves better results; nevertheless, the improvements are marginal. We attribute these minimal improvements to the quality of synthetic data. Following our previous study \cite{singh2022enhancing}, we only incorporated these datasets to improve the quality of the overall ASR model.

\subsection{Effectiveness of gradual filtration}
Figure \ref{fig:wer-cs} illustrates the performance of the model with different CS thresholds and the amount of PL data selected with filtration over multiple PL iterations. Initially, we start with aggressive filtering with (CS$\ge$0.5), which selects only 1.5\% of the total unlabeled Audiobooks dataset. With this threshold, although we select high-quality PL, however, it results in marginal improvements overall (2.8\% of relative improvement). As we gradually relax the confidence score threshold, our model indeed continues to produce better results over each iteration of PL. With (CS$\ge$0), selects 9.37\% of PL, which results in 9.5\% of overall relative improvement compared to baseline. Overall, as Figure \ref{fig:wer-cs} shows, (CS$\ge$-1.5), which selects 72.66\% of PL, produces the lowest WERs across most datasets. Most consistent improvements are seen on the Common Voice and Punjabi Speech datasets.       
Further, our experimental analysis reveals that PL generated with (CS$\ge$-2 or greater) selects more than 80\% of the total PL data. However, the performance of the model gradually starts to diminish as compared to the Best PL. 
\vspace{-0.3cm}
\section{Conclusions}
 The paper presents a self-training (pseudo-labeling) approach for automatic speech recognition for low-resource settings, specifically focusing on the Punjabi language.  The proposed self-training approach generates highly accurate pseudo-labels for unlabeled Punjabi speech, resulting in a significant improvement in word error rate (WER) compared to a strong baseline. The experimental analysis demonstrates the effectiveness of our approach and highlights the potential to improve the performance of ASR systems for low-resource languages such as Punjabi.

\section{Acknowledgement}
This work is supported by the 2020 Catalyst: Strategic New Zealand - Singapore Data Science Research Programme Fund, Ministry of Business, Innovation, and Employment (MBIE), New Zealand. Ruili Wang is the corresponding author. 
\bibliographystyle{IEEEtran}
\bibliography{main}

\end{document}